# An Effective AHP based Metaheuristic Approach to Solve Supplier Selection Problem

Tamal Ghosh[1], Tanmoy Chakraborty, Pranab K Dan
Department of Industrial Engineering & Management,
West Bengal University of Technology.
BF 142, Sector 1, Salt Lake City, Kolkata 700064, India.
Email: tamal.31@gmail.com, chakrabortytanmoy.85@gmail.com, danpk.wbut@gmail.com.

## Abstract

The supplier selection problem is based on electing the best supplier from a group of pre-specified candidates, is identified as a Multi Criteria Decision Making (MCDM), is proportionately significant in terms of qualitative and quantitative attributes. It is a fundamental issue to achieve a trade-off between such quantifiable and unquantifiable attributes with an aim to accomplish the best solution to the abovementioned problem. This article portrays a metaheuristic based optimization model to solve this NP-Complete problem. Initially the Analytic Hierarchy Process (AHP) is implemented to generate an initial feasible solution of the problem. Thereafter a Simulated Annealing (SA) algorithm is exploited to improve the quality of the obtained solution. The Taguchi robust design method is exploited to solve the critical issues on the subject of the parameter selection of the SA technique. In order to verify the proposed methodology the numerical results are demonstrated based on tangible industry data.

**Keywords:** supplier rating; vendor section; AHP; simulated annealing; metaheuristic; Taguchi method.

**1. Introduction**

Over the decades the supplier selection problem has been dragging the attention of the researchers and practitioners in the vicinity of Supply Chain Management study. Therefore several techniques have been proposed by the academicians and professionals to solve this Multi Criteria Decision Making (MCDM) problem. As Dickson (1966) stated that contemporary firms have the alternatives to select a suitable technique to decipher the complicated job of selecting suppliers. These techniques varies from the adoption of uncomplicated methods such as the selection of the vendor offering lowermost tender to the effective complicated methods which exploit the unquantifiable attributes of the supplier selection problem. Although many empirical studies are already contributed in the proposed area of research, yet the researchers and practitioners are involved in improving the solution methodologies due to the involvement of growing complexities of technology innovation. In manufacturing industry the procurement cost of raw materials from the outside suppliers and the other unquantifiable attributes such as quality of the materials delivered and the time of delivery are substantially crucial and nearly 70% to 80% of the total cost involved in the whole product development phase (Weber,

---

[1] Corresponding Author: Email: tamal.31@gmail.com, Phone: +91-33-2334-1014, Fax: +91-33-2334-1031.

Current and Benton, 1991). Since the early 1990's a set of hard to estimate unquantifiable attributes have turned out to be the key components of the supplier selection problem which further helped in developing novel and complicated methodologies (Ellram, 1990). Selecting an optimal set of suppliers on this basis certainly increase the sustainability of the firms in present global competition (Thompson, 1990). Traditionally the supplier selection problem is based on the purchasing cost, quality and delivery functionality of the raw materials. Consideration of the above aspects enhances the complexities of the problem in polynomial time. The selection decision becomes extensively significant for any person-in-charge of the procurement department due to the several different levels of success of the rival suppliers under these stated circumstances (Aissaoui, Haouari and Hassini, 2007). The vendor with the lowermost tender might not be the best in delivering the material of proper quality. In case of the government bodies, the practice of decision-making optimization model in vendor selection processes would be extremely helpful than the private firms. In the government framework, there is a legal need to obey the formal rules and procedures that control vendor selection processes. Therefore such decision models could be immensely supportive in maintaining transparency and doing fair business (Schooner, 2003). It is evident from the past literature that the precise and sophisticated methodologies are urgently required to solve the supplier selection problems (Scott, 1995). For that matter Kanagaraj and Jawahar (2009) have implemented a Simulated Annealing Algorithm (SAA) tool for supplier selection problems to obtain the optimal or near-optimal solutions quickly. Saen (2009) introduced a technique based on Data Envelopment Analysis (DEA) to rank the suppliers in the presence of nondiscretionary factors. An integrated model of Fuzzy-AHP and decision support system was proposed by Gnanasekaran, Velappan and Manimaran (2010) for supplier selection to reduce the cost and enhance the product quality with the help of the Reliability-Based Total Cost of Ownership (RBTCO) model. This integrates purchasing, maintenance and stoppage costs along with the realistic constraints based on product reliability and weight restraint. Nonlinear Integer Programming (NLIP) is used to develop the mathematical formulation of the RBTCO model.

In present article a multiobjective supplier selection problem has been introduced which has incorporated faulty materials and delay in delivery cost as decisive factors. Thereafter a Simulated Annealing (SA) based optimization model is implemented to solve this problem. The initial solution to the SA procedure is generated using Analytic Hierarchy Process (AHP). In order to select the optimal set of parameters to the SA, Taguchi's Design of Experiments (DOE) technique is exploited.

The rest of the article is structured in following manner, section 2 presents a brief literature survey of the proposed area of study, section 3 demonstrates the formulation of the proposed multiobjective problem model, section 4 elaborates the proposed solution methodology and numerical results are depicted in section 5 followed by the managerial implications and conclusion of this research.

## 2. Literature Survey

Li, Fun and Hung (1997) reported that the supplier selection methodologies are practiced to facilitate the selection process which further have a substantial impact on the selection outcomes. Numerous supplier selection methods have been established and categorized over the decades. Petroni and Braglia (2000) proposed linear weighting method using supplier rating based on different attributes which is a faster and inexpensive method to instigate. Although various drawbacks and limitations are also indicated in their study. Cost proportion (Timmerman, 1986) and aggregate cost of proprietorship (Ellram, 1990) based on aggregate cost approaches assemble all the cost components of the supplier selection process in fiscal units which are very flexible methods. These are exact methods and due to the complexities and time involvement these methods are moderately expensive to implement. Mathematical programming approaches primarily exploit the quantifiable factors; these approaches comprise the Principal Component Analysis (PCA) (Petroni and Braglia, 2000) and the Artificial Neural Network (ANN) (Choy, Lee and Low, 2002). Bello (2003) stated in his research that the PCA approach is advantageous in terms of its competency in managing various differing aspects. The ANN approaches are also useful in cost minimization and time reduction.

Multiple Attribute Utility Theory method is practiced for global vendor selection problems, in which the surroundings are more complex and uncertain (Zhao and Bross, 2004). Chen, Lin and Huang (2006) demonstrated that the Fuzzy set theory technique controls such situations for supplier performance evaluation. The approach is helpful to the managers to purchase from the suppliers according to their own choice. a fuzzy weighted additive and mixed integer linear programming is developed. The model aggregates weighted membership functions of objectives to construct the relevant decision functions, in which objectives have different relative importance. Ng (2008) proposed a weighted linear program for the supplier selection problem. This paper demonstrates a transformation technique with the mathematical model which can solve the problem without an optimizer. Amid, Ghodsypour and O'Brien (2009) demonstrated fuzzy multiobjective and additive model to consider the imprecision of information along with the order quantities to each supplier using price breaks. The objective functions used are, minimizing the net cost, rejected items and late deliveries, while satisfying other constraints such as capacity and demand requirement. Authors further stated weighted max–min model in the similar problem environment (Amid, Ghodsypour and O'Brien, 2010) with the help of Analytic Hierarchy Process (AHP). The proposed model could be utilized to find out the appropriate order to each supplier. Saen and Zohrehbandian (2008) proposed DEA approach with quantity discount policy to select the best supplier. Wu (2009) stated a hybrid model using data envelopment analysis (DEA), decision trees (DT) and neural networks (NNs) to evaluate supplier performance. The model comprised of two parts. The first part utilized DEA and categorizes the suppliers into several clusters thereafter the second part used the performance-related data of the company to train decision trees and intelligent neuro model, the precised results are obtained. However Saen (2008) stated an approach based on super-efficiency analysis based on DEA to rank the suppliers in presence of volume discounts. Author further proposed another DEA based method to rank the suppliers in the presence of weight limitations and dual-role factors (Saen, 2010). The Analytical Hierarchical Process (AHP) is the most exploited method in supplier selection problems, which is a Multi Criteria Decision Making technique. It is implemented to rank the alternatives when several criteria are believed to be considered and it permits the managers to formulate complicated problems in the form of a hierarchical relationship (Saaty, 1980). The AHP is comparatively straightforward method to practice. This technique integrates tangible and intangible attribute of the problems. A detail survey (Tahriri et al., 2008) of the supplier selection methods portrays that the AHP is the most frequently employed method in supplier selection. The AHP hierarchy typically contains three distinct levels, which are objectives, factors, and alternatives. AHP suggests an way to rank the alternative choices based on the manager's decisions relating the significance of the criteria. Due to this fact AHP is preferably appropriate for the abovementioned problem. The problem hierarchy provides itself to an analysis based on the impact of a given level on the next higher level (Saaty, 1980). Managerial judgments are stated in terms of the pair-wise comparisons of entries on a specified level of the hierarchy based on their influences on the next higher level. Each of the pair-wise comparisons signifies an approximation of the proportion of the weights of the two criteria being compared. Since AHP exploits a proportionate scale for personal decisions, the relative weights reflect the relative importance of the norms in attaining the objective of the hierarchy (Tam and Tummala, 2001). Yahya and Kingsman (1990) used Saaty's AHP method to determine primacy in selecting suppliers. The authors employed vendor rating in determining how to allot business and where inadequate progress work is utilized. Akarte (2001); Handfield, Walton and Sroufe (2002); Yu and Jing (2004); Liu and Hai (2005); Rajkumar, Kannan and Jayabalan (2009) also utilized AHP technique in their study as an integral part.

Most of the supplier selection literature generally used traditional methods. Traditional techniques are not efficient when the solution state space is large and various constraints cause the vendor selection problem more complicated. Very few articles utilized state-of-the-art methods such as metaheuristics. Recently Arunkumar, Karunamoorthy and Makeshwaraa, (2007); Rezaei and Davoodi (2008); Kubat, and Yuce, (2006); Ding, Benyoucef and Xie (2003) have proposed Genetic Algorithm based metaheuristic approach to solve supplier selection problems in multiobjective environment.
This present research introduces an efficient metaheuristic approach based on Simulated Annealing to solve the supplier selection problem.

## 3. Problem Formulation

Supplier selection processes are primarily reliant on the specific objectives being solved with the problem and the relevant constraints related to the objectives. In this article a case of a Leading construction firm of India has been considered to derive the multiobjective optimization model. This firm constructs commercial buildings as well as residential units in large scale such as IT/ITeS SEZ, Shopping malls, hospitality and retail units, logistics and industrial squares etc. Due to the heavy construction approach, this company requires various raw materials such as steel beams, cement, light weight bricks, cast iron etc.
The proposed model is articulated considering the constrictions which are managed rationally by this firm to select suppliers. Each raw material supplied by the supplier would have different constraints and characteristics such as percentage of faulty materials supplied, percentage delay in delivery and unit purchasing cost of the materials. Each supplier certainly has its own capacity to supply. The firm has specific requirements of material in certain period. If raw material $j$ is being supplied by supplier $k$, then the initial procurement cost is defined as,

$$P_c = \sum_{j=1}^{M} \sum_{k=1}^{N} \varphi_{jk} q_{jk} \qquad (1)$$

where
$\varphi_{jk}$ is the unit cost of $j^{th}$ material supplied by $k^{th}$ supplier.
$q_{jk}$ is the amount of material type $j$ procured by $i^{th}$ supplier.
The total substandard delivery is defined as,

$$Q_d = \sum_{j=1}^{M} \sum_{k=1}^{N} \delta_{jk} q_{jk} \qquad (2)$$

where
$\delta_{jk}$ is the percentage of faulty items of $j^{th}$ material supplied by $k^{th}$ supplier.
$q_{jk}$ is the amount of material type $j$ procured by $i^{th}$ supplier.

The total delay in delivery is defined as,

$$D_d = \sum_{j=1}^{M} \sum_{k=1}^{N} \theta_{jk} q_{jk} \qquad (3)$$

where
$\theta_{jk}$ is the delay percentage in delivery of $j^{th}$ material supplied by $k^{th}$ supplier.
$q_{jk}$ is the amount of material type $j$, procured by $i^{th}$ supplier.

In this supplier selection problem three objectives are simultaneously considered, (i) to minimize the total procurement cost, (ii) to minimize the total number of faulty items supplied and (iii) to minimize the total number of delay days in delivery in procuring various raw materials from various suppliers. The quality function (2) demonstrates the number of faulty items supplied by the suppliers. The faulty items are generally detected by the receiving firm while relocating the raw materials to the inventory. The firm's strategy is to return back the substandard items to the suppliers and request them to replace those items within stipulated time which is substantially one week. Therefore in a project under fixed schedule, supply of substandard material could incur the total cost in terms of one week delay time. Thus the quality function of equation (2) and total delay in delivery of equation (3) could be transformed into a non-compliance cost component using,

$$Q_c = (7 + D_d) \times C_d \qquad (4)$$

Hence the proposed multiobjective model aims in minimizing the total cost function $f(x)$,

$$Minimize\ f(x) = P_c + Q_c \qquad (5)$$

subject to

$$q_{jk} \leq X_{jk} \qquad (6)$$

$$\sum_{k=1}^{N} q_{jk} \geq Y_j \qquad (7)$$

where
$X_{jk}$ is the capacity of supplier $k$ while supplying material $j$
$Y_{jk}$ is the demand of material $j$ to the firm in certain period

Constraint (6) confirms that each supplier supplies according to its capacity. Constraint (7) confirms that total raw material procured should harmonize the firm's demand. This proposed multiobjective model is validated using simulated annealing (SA) approach by considering these abovementioned objectives and constraints which further gives the optimal selection results of suppliers. In order to define the SA approach, a quick solution generation method is believed to be identified and the generated feasible solution is assumed to be used as the initial solution to the SA algorithm. Therefore in this research Analytic Hierarchy Process (AHP) is utilized to serve the purpose. The next section would demonstrate the details of the solution methodologies involved in this approach.

## 4. Research Methodology

To facilitate the present research work authors have visited an eminent construction firm operating in Kolkata, India. Currently the firm is involved in a big project based on development of an IT/ITeS Special Economic Zone (SEZ) in sector V, Rajarhat, Newtown in Salt Lake City Kolkata. Authors have prepared the questionnaires based on the information gathered from the professionals of the abovementioned firm. On the basis of the experts opinion the AHP analysis has been carried out in this study.

The proposed optimization methodology is developed using Simulated Annealing (SA) algorithm to achieve the optimal solutions. The SA algorithm simulates the natural annealing process, where particles of a solid organize themselves into a thermal equilibrium. An introduction to SA can be obtained from the book by Aarts and Korst (1990). The general applications concerns combinatorial optimization problems of the following form where *S* is a predetermined set of feasible solutions.

$$\min_{x \in S} g(x) \qquad (8)$$

The algorithm exploits a pre-defined neighbourhood structure on '*S*'. A control parameter called temperature in resemblance to the natural annealing process governs the search behaviour. In each iteration, a neighbour solution *y* to the current solution *x* is figured out. If *y* has an improved objective function value than *x*, the solution *y* is accepted, that is, the current solution *x* is swapped by *y*. Alternatively if *y* does not attain a better objective function value than *x*, the solution *y* is only recognized with a specific probability depending on (i) the difference of the objective function values in *x* and *y*, and (ii) the temperature parameter. The pseudocode below exhibits the general SA method.

```
Pseudocode (SA)
initialize;
repeat
generate a candidate solution;
evaluate the candidate;
determine the current solution;
reduce the temperature;
until termination condition is met;
```

In this article, factors which affect the decision makers' strength are analysed. The process follows these steps,
- ➢ Using the AHP method, the weight of the factors are obtained from qualitative expressions and each supplier's weight is also achieved and final composite criteria weights are determined by AHP. This further rank the suppliers according to their composite scores. Therefore according to the ranks the suppliers could be selected.
- ➢ Simulated Annealing (SA) takes the ranks obtained from previous step and attempts to search for optimal set of suppliers based on the objectives defined in section 3.

*4.1. Initial Solution Generation*

In this article AHP is exploited to select initial set of suppliers. This solution is an initial feasible solution to the SA method based optimization model.

*4.2. Analytical Hierarchy Process (AHP)*

The AHP is a multi-criteria decision technique that exploits hierarchical relationships to represent a problem. Primacies for substitutes are acquired based on the opinion of the experts (Saaty, 1980). The method consists of several important steps: outlining the shapeless problem into shape, obtaining the AHP hierarchical relationships, forming pairwise comparison matrices, approximating the relative weights, examining the consistency and finally attaining the overall ranking (Lee, Chen and Chang, 2008). AHP can empower managers to represent the interface of several factors in complicated and shapeless circumstances. The technique is based on the pairwise comparison of decision variables with respect to the factors or substitutes. A pairwise comparison matrix is obtained of size $n \times n$, where $n$ is the number of criteria to be compared. The AHP method is adopted in this article is stated as,

Step 1: The hierarchical relationship of the problem is obtained and presented in Figure 1. The proposed AHP method decomposes the problem into three levels (Saaty, 1980): (i)The first level demonstrates the main objective: the selection of suppliers, the second level depicts the criteria and the last level reports the six suppliers to be compared.

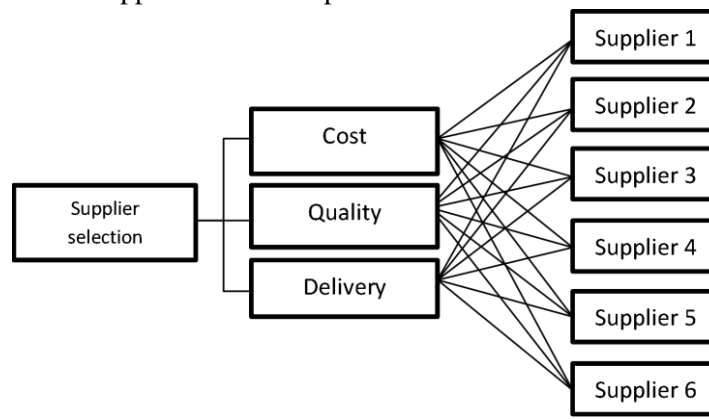

Figure 1. Hierarchical relationship of the supplier selection problem

Step 2: Calculation of the pairwise comparison matrix for each level is required. For the pairwise comparison, a ranking scale is used for the criteria evaluation (Saaty, 1980). The scale is a crisp scale ranging from 1 to 9, as presented in Table 1. This scale values are assigned to the criteria based on the experts opinion of the AHP questionnaire sheet prepared by the authors. The pairwise comparison matrix for all the criteria is presented in Table 2. The last column of table 2 depicts the overall importance of the criteria over each other.

The AHP procedure is presented as,
Consider $[Ax = \lambda_{max}x]$ where
$A$ is the comparison matrix of size $n \times n$, for $n$ criteria.
$x$ is the Eigenvector of size $n \times 1$
$\lambda_{max}$ is the Eigenvalue, $\lambda_{max} \in \mathcal{R} > n$.
To find the ranking of priorities, namely the Eigen Vector $X$:

1.  Initialization:
1.1. Take the squared power of matrix $A$, i.e., $A^2 = A \times A$
1.2. Find the row sums of $A^2$ and normalize this array to find $E_0$.
1.3. Set $A := A^2$

2. Main:
2.1. Take the squared power of matrix $A$, i.e., $A^2 = A \times A$
2.2. Find the row sums of $A^2$ and normalize this array to find $E_1$.
2.3. Find $D = E_1 - E_0$.
2.4. If the elements of $D$ are close to zero, then $X = E_1$
2.5. else set $A := A^2$, set $E_0 := E_1$ and go to Step 2.1.
2.6. STOP.

Table 1. Saaty's 9 point scale

| Importance Value | Definition | Description |
|---|---|---|
| 1 | Equal Strong | Two factors are equally Contributing to objective |
| 3 | Moderate Strong | One factor is marginally superior over other |
| 5 | Fairly Strong | One factor is strongly superior over other |
| 7 | Very Strong | One factor is very stongly superior over other |
| 9 | Absolute Strong | The highest level of superiority of one factor over other |
| 2, 4, 6, 8 | Intermediate Values | According to the negotiation required |

Table 2. Comparison matrix of main attributes

|  | Cost | Quality | Delivery | Relative Weight (PV) |
|---|---|---|---|---|
| **Cost** | 1 | 1/5 | 1/9 | 0.062941 |
| **Quality** | 5 | 1 | 1/3 | 0.265433 |
| **Delivery** | 9 | 3 | 1 | 0.671625 |

Table 3 to Table 5 present the pairwise comparison matrices of the suppliers with each of the criteria selected in this research. The last columns of pairwise matrices present the calculated relative weights of the suppliers over each other.

Table 3. Pairwise comparison matrix of suppliers with respect to cost

|  | $S_1$ | $S_2$ | $S_3$ | $S_4$ | $S_5$ | $S_6$ | Relative Weight |
|---|---|---|---|---|---|---|---|
| $S_1$ | 1 | 3 | 2 | 7 | 6 | 5 | 0.40201 |
| $S_2$ | 1/3 | 1 | 1/2 | 5 | 4 | 3 | 0.18549 |
| $S_3$ | 1/2 | 2 | 1 | 4 | 3 | 2 | 0.21024 |
| $S_4$ | 1/7 | 1/5 | 1/3 | 1 | 1/2 | 1/4 | 0.03831 |
| $S_5$ | 1/6 | 1/4 | 1/3 | 2 | 1 | 1/3 | 0.05611 |
| $S_6$ | 1/5 | 1/3 | 1/2 | 4 | 3 | 1 | 0.10786 |

Table 4. Pairwise comparison matrix of suppliers with respect to quality

|  | $S_1$ | $S_2$ | $S_3$ | $S_4$ | $S_5$ | $S_6$ | Relative Weight |
|---|---|---|---|---|---|---|---|
| $S_1$ | 1 | 3 | 5 | 1/3 | 7 | 8 | 0.26949 |
| $S_2$ | 1/3 | 1 | 4 | 1/4 | 5 | 7 | 0.15914 |
| $S_3$ | 1/5 | 1/4 | 1 | 1/5 | 3 | 4 | 0.07354 |
| $S_4$ | 3 | 4 | 5 | 1 | 8 | 9 | 0.43202 |
| $S_5$ | 1/7 | 1/5 | 1/3 | 1/8 | 1 | 3 | 0.04087 |
| $S_6$ | 1/8 | 1/7 | 1/4 | 1/9 | 1/3 | 1 | 0.02493 |

Table 5. Pairwise comparison matrix of suppliers with respect to delivery

|     | $S_1$ | $S_2$ | $S_3$ | $S_4$ | $S_5$ | $S_6$ | Relative Weight |
|-----|-----|-----|-----|-----|-----|-----|-----------------|
| $S_1$ | 1   | 5   | 2   | 1/3 | 6   | 7   | O.26324         |
| $S_2$ | 1/5 | 1   | 2   | 1/4 | 3   | 4   | 0.11631         |
| $S_3$ | 1/2 | 1/2 | 1   | 1/5 | 3   | 5   | 0.10426         |
| $S_4$ | 3   | 4   | 5   | 1   | 7   | 9   | 0.43627         |
| $S_5$ | 1/6 | 1/3 | 1/3 | 1/7 | 1   | 4   | 0.05268         |
| $S_6$ | 1/7 | 1/4 | 1/5 | 1/9 | 1/4 | 1   | 0.02723         |

The overall rating of each supplier is computed by adding the product of the relative weight of each criterion and the relative weights of the suppliers considering the corresponding criteria (Table 6). Table 6 demonstrates that supplier 4 (overall ranking is 1 and composite score is 0.4099) is the best supplier followed by supplier 1, 2, 3, 5 and 6.

*4.3 Computation of consistency index and consistency ratio*

AHP procedure requires the computation of consistency ratio to ensure the precision of the obtained solution. The consistency index (*CI*) of pairwise comparison matrix is calculated using,

$$CI = \frac{\lambda_{max} - n}{n - 1} \tag{9}$$

and the random consistency index (*RI*) is computed as,

$$RI = 1.98 \frac{n - 2}{n} \tag{10}$$

where $\lambda_{max}$ is the maximum eigenvalue and *n* is the size of pairwise comparison matrix. Thus the consistency ratio (*CR*) is obtained using,

$$CR = \frac{CI}{RI} \tag{11}$$

*CI* and *CR* are computed to understand the consistency of the solution obtained. In general CI and CR are believed to be < 0.1. Table 7 provides the computed values of the CI and CR for all the pairwise comparison matrices. All the computed values are < 0.1. Therefore the computed results are acceptable.

Table 6. Composite score matrix from above four matrices

|     | Cost (PV=0.0629) | Quality (PV=0.2654) | Delivery (PV=0.6716) | Composite Score (Rank) |
|-----|------------------|---------------------|----------------------|------------------------|
| $S_1$ | 0.402014         | 0.26949             | 0.263243             | **0.2733 (2)**         |
| $S_2$ | 0.18549          | 0.159145            | 0.11631              | **0.1319 (3)**         |
| $S_3$ | 0.210246         | 0.0735405           | 0.104265             | **0.1026 (4)**         |
| $S_4$ | 0.0383076        | 0.432023            | 0.436267             | **0.4099 (1)**         |
| $S_5$ | 0.0561086        | 0.0408689           | 0.0526848            | **0.0496 (5)**         |
| $S_6$ | 0.107865         | 0.0249321           | 0.0272297            | **0.0315 (6)**         |

Table 7. CI, RI, CR values for all the pairwise comparison matrices

|     | Cost     | Quality   | Delivery | Between criteria |
|-----|----------|-----------|----------|------------------|
| **CI** | 0.055157 | 0.0944284 | 0.095729 | 0.0145319        |
| **RI** | 1.32     | 1.32      | 1.32     | 0.66             |
| **CR** | 0.0417   | 0.07153   | 0.07252  | 0.022            |

*4.4. Fitness function of SA procedure*

SA procedure examines the fitness score of each solution generated in the solution neighbourhood. Fitness calculation is one of the most significant steps of the metaheuristic method because it decides which solution is to be stored and which one is to be eliminated. The multiobjective function *f(x)* is required to compute the fitness score. In order to facilitate the computation, the *f(x)* is transformed into *F(x)* and expressed in equation (12). Two weights *weight1* and *weight2* are also assigned in the equation.
It is considered that *weight1+weight2=1*.

$$F(x) = \frac{total\ cost}{total\ cost + quality\ cost} \times weight1 + \frac{total\ cost}{total\ cost + delay\ cost} \times weight2 \qquad (12)$$

where total cost is computed by summing up total procurement cost and non-compliance cost. Since the firm gives more importance to the delivery issues rather than the quality or cost issues, thus *Weight1* and *weight2*, are assigned with prefixed values of 0.3 and 0.7 respectively.

*The Proposed Simulated Annealing Algorithm*

This subsection describes the proposed SA algorithm in depth. The initial input is a solution string which is generated from AHP technique. Therefore the initial input string $S_0$ which is obtained from Table 6 using the ranks of the suppliers,

$$S_0 = [2, 3, 4, 1, 5, 6]$$

The size of the solution string is 6 which is the total number of suppliers to be evaluated. Each bit of the string represents the rank of the corresponding supplier (string index). Therefore $S_0$ states supplier 1 to 6 retain the ranks 2, 3, 4, 1, 5 and 6 respectively.
Thereafter this multiobjective SA procedure is set to maximize *F(x)* of equation (12). Some symbolizations used in the algorithm are introduced as,

$S_{cur}$ → *current solution*
$S_i$ → *neighbourhood solution*
$S_{best}$ → *best solution found so far*
$T_{init}$ → *initial temperature*
$T_{final}$ → *freezing temperature*
$T$ → *current temperature*
$α$ → *temperature reducing factor*
$M$ → *Markov chain length*
*iter* → *iteration number*
$f_i$ → *current fitness value*
$f_{best}$ → *best fitness value*

The steps of the proposed algorithm can be summarized as follows.

*Step 1. Obtain an initial solution $S_0$ by using Analytic Hierarchy Process method*
*Step 2. Evaluate $S_0$ and Calculate corresponding fitness value $F_0$; $F_0 = F(S_0)$)*
*Step 3.* **Set $F_{best}= F_0$, Set** *$S_{best} = S_0 = S_{cur}$.*
*Step 4. Initialize SA Heuristic and its parameters: $T_{init}$, $T_{final}$, α, M, iter = 0, count=0, count1=0.*
*Step 5.* **If** *count < M, then repeat Steps 5.1 to 5.9.*
*Step 5.1. Generate a new supplier rank configuration neighbourhood searching by performing exchange-move (randomly selecting two suppliers and interchanging their ranks).*
*Step 5.2. Read suppliers rank configuration from above steps and generate corresponding neighbourhood solution $S_i$.*
*Step 5.3.* **If** *$F(S_i) > F_{best}$, then $S_{best} = S_i$, $S_{cur} = S_i$, count = count + 1,* **go to** *Step 5.*

*Step 5.4.* **If** *F(S$_i$) = F$_{best}$,* then *S = S$_i$, count1= count1 + 1, count = count + 1,* **go to** *Step 5.*
*Step 5.5.* Compute *δ = F(S$_i$)- F(S$_{cur}$).* Obtain a random variable *r* in the range of *U(0,1).*
*Step 5.6.* **If** *e$^{δ/T}$ > r,*
*Step 5.7.  set S$_{cur}$ = S$_i$,*
*Step 5.8. count1= 0;*
*Step 5.9.* **else** *count1 = count1 + 1.*
*Step 5.10. iter = iter + 1.*
*Step 5.11.* **until** *freezing temperature (T$_{final}$) is reached;*
*Step 5.12. reduce the temperature using T$_i$ = α×T$_{i-1}$ function;*

The SA procedure is repetitively employed until a solution is achieved which attains the highest fitness score. All the parameters and counters are initialized in step 4. A special move, namely exchange-move, is utilized in the proposed algorithm to guide the solution searching procedure. It is spotted that exchange-move ordinarily leads to the improved solutions effortlessly and competently which is practiced as a principle component for finding better neighbourhood solution in step 5.1. The algorithm also verifies the number of instances when neighbourhood solutions become static. If this number attains a pre-fixed constant value, the fitness value of current configuration is compared to the optimal solution obtained thus far to conclude whether to prolong the iterations or stop with the best solution achieved.

**5. Experiments and Verifications**

In order to apply the proposed SA algorithm as a solution methodology the effects of changing the values of the various parameters are studied. Determining the optimal set of parameters are crucial in this respect. Therefore in this article Taguchi's robust design method (Taguchi, 1994) is employed to determine the optimal parameters set.

*5.1 Taguchi Method for Parameters Selection*

The parameters are Initial temperature ($T_{init}$), temperature reducing factor (*α*) and Markov chain length (*M*) (final temperature ($T_{final}$) is taken as constant value = -∞) and termed as factors, and each factor has three discrete levels (Table 8). Hence an *L9* orthogonal array is used, and this recommends that 9 sets of Taguchi experiments are prerequisite and the results are evaluated by using an analysis of variance (ANOVA) technique. The parameter settings for each experiment are shown in Table 9.

Table 8. Levels of parameters tested

| levels | Parameters | | |
|---|---|---|---|
| | *M* | *α* | *T$_{init}$* |
| 1 | 20 | 0.75 | 10 |
| 2 | 30 | 0.85 | 20 |
| 3 | 40 | 0.95 | 30 |

Table 10 presents the results of the corresponding ANOVA analysis with Signal-to-Noise ratio (Larger-the-better). In Table 10, the variance ratios (*F* ratios) of the factors are determined. A test of significance at 95% confidence level is employed to spot the significance of these factors. The *P* values of the factors $T_{init}$, *α, M* are investigated and all the values of the parameters are seen to be less than the critical level with degrees of freedom at (2, 8). This suggests that all the parameters are significant factors in the proposed approach. The response table (Table 11) depicts the average of each response characteristic for each level of each of the factors. Table 11 include the ranks based on Delta (*δ*) statistic, which compares the relative magnitude of effects. Ranks are assigned based on *δ* values. Using the level averages in the response table optimal set of levels of the factors could be determined which yields the best result.

Table 9. The Experimental Settings of the Taguchi Experiments

| Experiments | $T_{init}$ | α | M | responses |
|---|---|---|---|---|
| 1 | 10 | 0.75 | 20 | 0.962364 |
| 2 | 10 | 0.85 | 30 | 0.923400 |
| 3 | 10 | 0.95 | 40 | 0.945652 |
| 4 | 20 | 0.75 | 30 | 0.956225 |
| 5 | 20 | 0.85 | 40 | 0.945652 |
| 6 | 20 | 0.95 | 20 | 0.932652 |
| 7 | 30 | 0.75 | 40 | 0.947348 |
| 8 | 30 | 0.85 | 20 | 0.962652 |
| 9 | 30 | 0.95 | 30 | 0.959657 |

Table 10. ANOVA table

| Factors | Degrees of Freedom | Factor Sum of squares | Mean Square (Variance) | F Ratio | P Value |
|---|---|---|---|---|---|
| $T_{init}$ | 2 | 0.000301 | 0.000150 | 0.34 | 0.744 |
| α | 2 | 0.000222 | 0.000111 | 0.25 | 0.798 |
| M | 2 | 0.000078 | 0.000039 | 0.09 | 0.918 |
| Residual Error | 2 | 0.000875 | 0.000438 | | |
| Total | 8 | 0.001475 | | | |

The ranks indicate that Initial temperature ($T_{init}$) has the greatest influence followed by temperature reducing factor (α) and Markov chain length (M). Each factor level should be fixed in such a way that the highest response could be achieved. Table 11 and the main effects plot of Figure 2 show that the optimal solution is obtained when $T_{init}$, α and M are set to 30, 0.75 and 20 respectively.

Table 11. Response table

| Levels | $T_{init}$ | α | M |
|---|---|---|---|
| 1 | 0.9438 | 0.9553 | 0.9526 |
| 2 | 0.9448 | 0.9439 | 0.9464 |
| 3 | 0.9566 | 0.9460 | 0.9462 |
| δ | 0.0127 | 0.0114 | 0.0063 |
| Rank | 1 | 2 | 3 |

*5.2 Convergence Analysis*

Convergence analysis of the SA procedure is quite simple for the supplier selection problem. The convergence curve during iterations of the proposed metaheuristic technique is presented in Figure 3. For the first iteration the fitness score attained a value of 0.9622. Since the SA procedure is designed to maximize the fitness function with the iteration counts therefore at 4[th] iteration it attained the value of 0.9633, an increase of 0.1% which is the final optimal solution. Based on the experimentation reported in this article, it is observed that the fitness score is increased with the iteration counts till it reaches the best fitness score at some iteration and thereafter the fitness score continues to remain constant even if the number of iterations is increased. Therefore the convergence property is established. For the test problem in hand the proposed approach is executed for 63 iterations and took 1.9281 CPU seconds to attain the best solution which proves its computational efficiency.

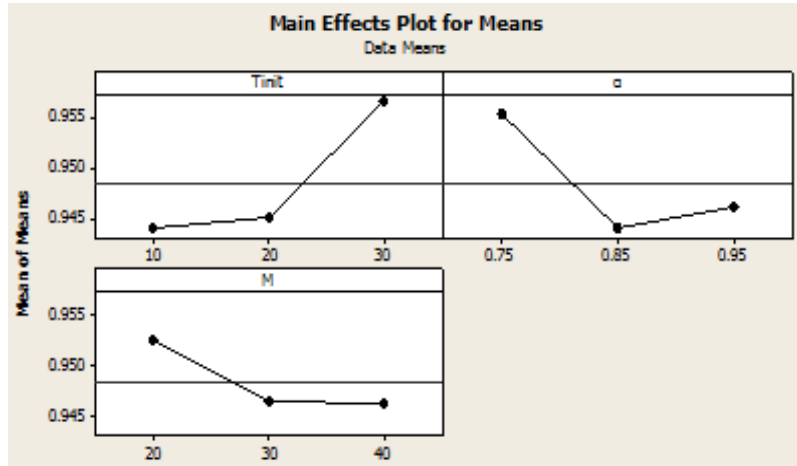

Figure 2. Main effects plot

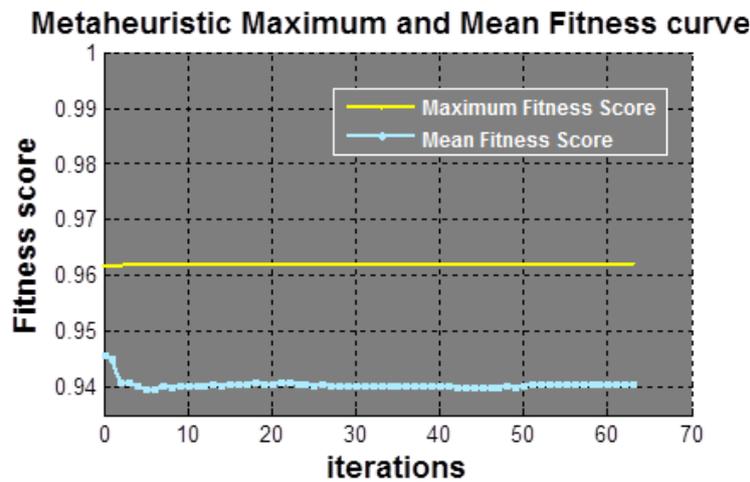

Figure 3. Convergence curve of SA algorithm

*5.3 Computational Results*

Data for six vendors are collected from the construction firm for a periodic demand of 600 metric ton TMT steel bar and depicted in Table 12. The strategy of the procurement department is to distribute the order among the best three vendors to avoid the biasness. Each supplier has certain capacity to supply materials as given in table 12. Each supplier supply faulty materials according to the percentage of defective items calculated and each of them supply according to their own pace and the percentages of delays in delivery are also provided in the table 12 for each supplier. The firm's project schedule is moderately rigid and therefore delay in delivery incurs the overall cost of the project. Delay cost is calculated by the experts of the firm, which is closely 2.5 Lacs INR per day. The cost incurs due to substandard supply are generally converted into delay cost as stated by the firm's manager and calculated using the first component of equation (4).

The AHP method depicts that supplier 4, 1, 2 are the best three suppliers. The total procurement cost is obtained for these three vendors are outlined in Table 13. Thereafter the SA procedure is executed and different result is obtained which states that supplier 3, 1, 4 are the best. The total procurement cost computed for SA method is found to be less than the AHP result and shown in Table 13. Although total faulty materials supplied and total delay days are almost identical for both the methods but the SA method attains closely 0.3% better solution than that of the AHP method. In monetary term SA recovers 1.25 Lacs INR for the firm. This observation indicates that the SA technique is efficient and less complex because of its simplicity in simulation. The solution is obtained with negligible computational time (< 2 seconds). Thus the proposed SA method is shown to outperform the AHP technique.

Table 12. Collected vendor details data from the firm

| Vendors | max order quantity (metric ton) | unit cost (INR)/kg | Percent defective/metric ton | percent delay delivery |
|---|---|---|---|---|
| v1 | 150 | 58.75 | 3.2 | 2.28 |
| v2 | 300 | 62 | 3.8 | 2.92 |
| v3 | 250 | 61.5 | 4.5 | 3.12 |
| v4 | 200 | 65 | 2 | 1.16 |
| v5 | 750 | 64.5 | 5.7 | 3.29 |
| v6 | 450 | 63.5 | 6.2 | 4.5 |

Table 13. Comparison of SA result with AHP result

| | vendor | order | defective | late delivery | cost (INR) |
|---|---|---|---|---|---|
| **AHP result** | v4 | 200 | 4 | 3 | 13000000 |
| | v1 | 150 | 5 | 4 | 8812500 |
| | v2 | 250 | 10 | 8 | 15500000 |
| | | **600** | **19** | **15** | **37312500** |
| | | | | quality cost | 1750000 |
| | | | | delay cost | 3750000 |
| | | | | Total procurement Cost | **43812500** |

| | vendor | order | defective | late delivery | cost (INR) |
|---|---|---|---|---|---|
| **SA result** | v3 | 250 | 11 | 8 | 15375000 |
| | v1 | 150 | 5 | 4 | 8812500 |
| | v4 | 200 | 4 | 3 | 13000000 |
| | | **600** | **20** | **15** | **37187500** |
| | | | | quality cost | 1750000 |
| | | | | delay cost | 3750000 |
| | | | | Total procurement Cost | **43687500** |

## 6. Managerial Implications

The study accomplished in this research has significant managerial implications. The soft computing approach proposed in this article can be exploited as a critical managerial decision making tool. This is beneficial in optimizing the vendor network, successful resource allocation for vendor improvement curriculums.

In optimizing the vendor network, managers can employ the method to choose vendors without having any biasness for any particular vendor. This further reduces the chance of failure in supplier network. By doing this it would help every supplier to grow evenly. The management of the firm can deliver these suppliers with possible standards for enhancement and target time could be anticipated in complementing them.

Another managerial insight of this study affirms that, an already recognized supplier selection methodology (AHP) may not be the best methodology and other state-of-the-art techniques (metaheuristics) can substantially attain better solution and maximize the profit of the firm which is the firms' main objective.

## 7. Conclusions

This article portrays a novel SA based metaheuristic algorithm to select the supplier for a particular Indian firm which is an NP-complete problem in nature. The problem is formulated using multiobjective mathematical model which reflects the essential optimization criteria of this research. The initial feasible solution to the proposed SA based technique is obtained using AHP technique in order to quicken the computation. This work further exploits Taguchi's robust design approach to select optimal set of parameters to SA algorithm which is crucial in influencing the performance of the technique. The uniqueness of this work lies in practicing two different decision making techniques in solving this MCDM problem model. In past literature such metaheuristic approach to evaluate and enhance the AHP ranking of vendors has never been carried out. To perform the said analysis authors have collected industrial data from a national construction firm. Computational results presented in Section 5 demonstrate that the SA method outperforms the AHP technique performing better than the AHP method for the supplier selection problem. The proposed SA procedure produces nearly 0.3% improved solution. This work is an experimental study which considered the main criteria of the problem such as cost, delivery and delay. However many other intricate sub-criteria could also be considered to make this work more realistic. Future work can be accomplished by utilizing this technique in more complex supplier selection problems incorporating more conflicting criteria and sub-criteria by considering risk factors or suppliers profiles and other related issues and that is the possible extension of this research.

**Acknowledgement**

The authors are grateful to the anonymous reviewers for their valuable comments and suggestions in improving the quality of the paper.